# Robustness of Causal Claims

Judea Pearl
Cognitive Systems Laboratory
Computer Science Department
University of California, Los Angeles, CA 90024
*judea@cs.ucla.edu*

## Abstract

A causal claim is any assertion that invokes causal relationships between variables, for example, that a drug has a certain effect on preventing a disease. Causal claims are established through a combination of data and a set of causal assumptions called a "causal model." A claim is robust when it is insensitive to violations of some of the causal assumptions embodied in the model. This paper gives a formal definition of this notion of robustness, and establishes a graphical condition for quantifying the degree of robustness of a given causal claim. Algorithms for computing the degree of robustness are also presented.

## 1  INTRODUCTION

A major issue in causal modeling is the problem of assessing whether a conclusion derived from a given data and a given model is in fact correct, namely, whether the causal assumptions that support a given conclusion actually hold in the real world. Since such assumptions are based primarily on human judgment, it is important to formally assess to what degree the target conclusions are sensitive to those assumptions or, conversely, to what degree the conclusions are *robust* to violations of those assumptions.

This paper gives a formal characterization of this problem and reduces it to the (inverse of) the identification problem. In dealing with identification, we ask: are the model's assumptions sufficient for uniquely substantiating a given claim. In dealing with robustness (to model misspecification), we ask what conditions must hold in the real world before we can guarantee that a given conclusion, established from real data, is in fact correct. Our conclusion is then said to be robust to any assumption outside that set of conditions.

To solve the robustness problems, we need techniques for quickly verifying whether a given model permits the identification of the claim in question. Graphical methods were proven uniquely effective in performing such verification, and this paper generalizes these techniques to handle the problem of robustness.

Our analysis is presented in the context of linear models, where causal claims are simply functions of the parameters of the model. However, the concepts, definition and some of the methods are easily generalizable to non parametric models, where a claim is any function computable from a (fully specified) causal model.

Section 2 introduces terminology and basic definitions associated with the notion of model identification and demonstrates difficulties associated with the conventional definition of parameter over-identification. Section 3 demonstrates these difficulties in the context of a simple example. Section 4 resolves these difficulties by introducing a refined definition of over-identification in terms of "minimal assumption sets." Section 5 establish graphical conditions and algorithms for determining the degree of robustness (or, over-identification) of a causal parameter. Section 6 recasts the analysis in terms of the notion of relevance.

## 2  PRELIMINARIES: LINEAR MODELS AND PARAMETER IDENTIFICATION

A linear model $M$ is a set of linear equations with (zero or more) free parameters, $p, q, r, \ldots$, that is, unknown parameters whose values are to be estimated from a combination of assumptions and data. The assumptions embedded in such a model are of several kinds: (1) zero (or fixed) coefficients in some equations, (2) equality or inequality constraints among some of the parameters and (3) zero covariance relations among error terms (also called disturbances). Some of these assumptions are encoded implicitly in the equations (e.g., the absence of certain variables in an equation), while others are specified explicitly, using expressions such as: $p = q$ or $cov(e_i, e_j) = 0$.

An instantiation of a model $M$ is an assignment of values to the model's parameters; such instantiations will be denoted as $m_1, m_2$ etc. The value of parameter $p$ in instantiation $m_1$ of $M$ will be denoted as $p(m_1)$. Every instantiation $m_i$ of model $M$ gives rise to a unique covariance matrix $\sigma(m_i)$, where $\sigma$ is the population covariance matrix of the observed variables.

**Definition 1** *(Parameter identification)*
*A parameter $p$ in model $M$ is identified if for any two instantiations of $M$, $m_1$ and $m_2$, we have:*

$$p(m_1) = p(m_2) \text{ whenever } \sigma(m_1) = \sigma(m_2)$$

In other words, $p$ is uniquely determined by $\sigma$; two distinct values of $p$ imply two distinct values of $\sigma$, one of which must clash with observations.

**Definition 2** *(Model identification)*
*A model $M$ is identified iff all parameters of $M$ are identified.*

**Definition 3** *(Model over-identification and just-identification)*
*A model $M$ is over-identified if (1) $M$ is identified and (2) $M$ imposes some constraints on $\sigma$, that is, there exists a covariance matrix $\sigma'$ such that $\sigma(m_i) \neq \sigma'$ for every instantiation $m_i$ of $M$. $M$ is just-identified if it is identified and not over-identified, that is, for every $\sigma'$ we can find an instantiation $m_i$ such that $\sigma(m_i) = \sigma'$.*

Definition 3 highlights the desirable aspect of over-identification — *testability*. It is only by violating its implied constraints that we can falsify a model, and it is only by escaping the threat of such violation that a model attains our confidence, and we can then state that the model and some of its implications (or *claims*) are *corroborated* by the data.

Traditionally, model over-identification has rarely been determined by direct examination of the model's constraints but, rather indirectly, by attempting to solve for the model parameters and discovering parameters that can be expressed as two or more distinct[1] functions of $\sigma$, for example, $p = f_1(\sigma)$ and $p = f_2(\sigma)$. This immediately leads to a constraint $f_1(\sigma) = f_2(\sigma)$ which, according to Definition 3, renders the model over-identified, since every $\sigma'$ for which $f_1(\sigma') \neq f_2(\sigma')$ must be excluded by the model.

[1] Two functions $f_1(\sigma)$ and $f_2(\sigma)$ are distinct if there exists a $\sigma'$ such that $f_1(\sigma') \neq f_2(\sigma')$.

In most cases, however, researchers are not interested in corroborating the model in its entirety, but rather in a small set of claims that the model implies. For example, a researcher may be interested in the value of one single parameter, while ignoring the rest of the parameters as irrelevant. The question then emerges of finding an appropriate definition of "parameter over-identification," namely, a condition ensuring that the parameter estimated is corroborated by the data, and is not totally a product of the assumption embedded in the model.

This indirect method of determining model over-identification (hence model testability) has led to a similar method of labeling the *parameters* themselves as over-identified or just-identified; parameters that were found to have more than one solution were labeled over-identified, those that were not found to have more than one solution were labeled just-identified, and the model as a whole was classified according to its parameters. In the words of Bollen (1989, p. 90) "A model is over-identified when each parameter is identified and at least one parameter is over-identified. A model is exactly identified when each parameter is identified but none is over-identified."

Although no formal definition of parameter over-identification has been formulated in the literature, save for the informal requirement of having "more than one solution" [MacCallum, 1995, p. 28] or of being "determined from $\sigma$ in different ways" [Joreskog, 1979, p. 108], the idea that parameters themselves carry the desirable feature of being over-identified, and that this desirable feature may vary from parameter to parameter became deeply entrenched in the literature. Paralleling the desirability of over-identified models, most researchers expect over-identified parameters to be *more robust* than just-identified parameters. Typical of this expectation is the economists' search for two or more instrumental variables for a given parameter [Bowden and Turkington, 1984].

The intuition behind this expectation is compelling. Indeed, if two distinct sets of assumptions yields two methods of estimating a parameter and if the two estimates happen to coincide in data at hand, it stands to reason that the estimates are correct, or, at least robust to the assumptions themselves. This intuition is the guiding principle of this paper and, as we shall see, requires a careful definition before it can be applied formally.

If we take literally the criterion that a parameter is over-identified when it can be expressed as two or more distinct functions of the covariance matrix $\sigma$, we get the untenable conclusion that, if one parameter is over-identified, then every other (identified) parame-ter in the model must also be over-identified. Indeed, whenever an over-identified model induces a constraint $g(\sigma) = 0$, it also yields (at least) two solutions for any identified parameter $p = f(\sigma)$, because we can always obtain a second, distinct solution for $p$ by writing $p = f(\sigma) - g(\sigma)t(\sigma)$, with arbitrary $t(\sigma)$. Thus, to capture the intuition above, additional qualifications must be formulated to refine the notion of "two distinct functions." Such qualifications will be formulated in Section 4. But before delving into this formulation we present the difficulties in defining robustness (or over-identification) in the context of simple examples.

## 3 EXAMPLES

**Example 1** *Consider a structural model $M$ given by the chain in Figure 1,*

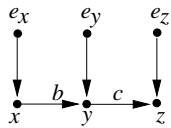

Figure 1:

*which stands for the equations:*

$$x = e_x$$
$$y = bx + e_y$$
$$z = cy + e_z$$

*together with the assumptions $cov(e_i, e_j) = 0$, $i \neq j$.[2] This model is identified because the model equations are regression equations; e.g., $E(ye_y) = 0$ and $E(ze_z) = 0$, hence $b = R_{yx}$ and $c = R_{zy}$, where $R_{yx}$ is the regression coefficient of $y$ on $x$.*

*Moreover, this model is over-identified, because it implies the conditional independence of $x$ and $z$, given $y$, which translates to the constraint: $R_{zx} = R_{yx}R_{zy}$.*

If we express the elements of $\sigma$ in terms of the structural parameters, we obtain:

$$R_{yx} = b$$
$$R_{zx} = bc$$
$$R_{zy} = c$$

where $R_{yx}$ is the regression coefficient of $y$ on $x$. $b$ and $c$ can each be derived in two different ways:

$$b = R_{yx} \quad b = R_{zx}/R_{zy} \qquad (1)$$

---

[2]Throughout this paper we assume recursivity and that the variables are correctly ordered. If non-recursive models are deemed feasible, additional assumptions need be stated explicitly, to rule out cycles, e.g., no arrow from $z$ to $y$.

and

$$c = R_{zy} \quad c = R_{zx}/R_{yx} \qquad (2)$$

which leads to the constraint

$$R_{zx} = R_{yx}R_{zy} \qquad (3)$$

If we take literally the criterion that a parameter is over-identified when it can be expressed as two or more distinct functions of the covariance matrix $\sigma$, we get the untenable conclusion that both $b$ and $c$ are over-identified. However, this conclusion clashes violently with intuition.

To see why, imagine a situation in which $z$ is not measured. The model reduces then to a single link $x \to y$, in which parameter $b$ can be derived in only one way, giving

$$b = R_{yx}$$

and $b$ would be classified as just-identified. In other words, the data does not corroborate the claim $b = R_{yx}$ because this claims depends critically on the untestable assumption $cov(e_x, e_y) = 0$ and there is nothing in the data to tell us when this assumption is violated.

The addition of variable $z$ to the model merely introduces a noisy measurement of $y$, and we can not allow a parameter ($b$) to turn over-identified (hence more robust) by simply adding a noisy measurement ($z$) to a precise measurement of $y$. We cannot gain any information (about $b$) from such measurement, once we have a precise measurement of $y$.

This argument cannot be applied to parameter $c$, because $x$ is not a noisy measurement of $y$, it is a cause of $y$. The capacity to measure new causes of a variable often leads to more robust estimation of causal parameters. (This is precisely the role of instrumental variables.) Thus we see that, despite the apparent symmetry between parameters $b$ and $c$, there is a basic difference between the two. $c$ is over-identified while $b$ is just-identified. Evidently, the two ways of deriving $b$ (Eq. 1) are not independent, while the two ways of deriving $c$ (Eq. 2) are.

Our next section makes this distinction formal.

## 4 ASSUMPTION-BASED OVER-IDENTIFICATION

**Definition 4** *(Parameter over-identification)*
*A parameter $p$ is over-identified if there are two or more distinct sets of logically independent assumptions in $M$ such that:*

1. each set is sufficient *for deriving the value of p as a function of* $\sigma$, $p = f(\sigma)$,

2. each set induces a distinct *function* $p = f(\sigma)$,

3. each assumption set is minimal, *that is, no proper subset of those assumptions is sufficient for the derivation of p.*

Definition 4 differs from the standard criterion in two important aspects. First, it interprets multiplicity of solutions in terms of distinct sets of assumptions underlying those solutions, rather than distinct functions from $\sigma$ to $p$. Second, Definition 4 insists on the sets of assumptions being minimal, thus ruling out redundant assumptions that do not contribute to the derivation of $p$.

**Definition 5** *(Degree of over-identification)*
*A parameter $p$ (of model $M$) is identified to degree $k$ (read: k-identified) if there are $k$ distinct sets of assumptions in $M$ that satisfy the conditions of Definition 4. $p$ is said to be m-corroborated if there are $m$ distinct sets of assumptions in $M$ that satisfy conditions (1) and (3) of Definition 4, possibly yielding $k < m$ distinct estimands for $p$.*

**Definition 6** *A parameter $p$ (of model $M$) is said to be* just-identified *if it is identified to the degree 1 (see Definition 5) that is, there is only one set of assumptions in $M$ that meets the conditions of Definition 4.*

Generalization to non-linear, non-Gaussian systems is straightforward. Parameters are replaced with "claims" and $\sigma$ is replaced with the density function over the observed variables.

We shall now apply Definition 4 to the example of Figure 1 and show that it classifies $b$ as just-identified and $c$ as over-identified. The complete list of assumptions in this model (assuming a known causal order) reads:

(1) $x = e_x$

(2) $y = bx + e_y$

(3) $z = cy + dx + e_z$

(4) $cov(e_z, e_x) = 0$

(5) $cov(e_z, e_y) = 0$

(6) $cov(e_x, e_y) = 0$

(7) $d = 0$

There are three distinct minimal sets of assumptions capable of yielding a unique solution for $c$; we will denote them by $A_1, A_2,$ and $A_3$.

**Assumption set $A_1$**

(1) $x = e_x$

(2) $y = bx + e_y$

(3) $z = cy + dx + e_z$

(5) $cov(e_z, e_y) = 0$

(6) $cov(e_x, e_y) = 0$

This set yields the estimand: $c = R_{zy \cdot x} = (R_{zy} - R_{zx}R_{yx})/(1 - R_{yx}^2)$,

**Assumption set $A_2$**

(1) $x = e_x$

(2) $y = bx + e_y$

(3) $z = cy + dx + e_z$

(4) $cov(e_z, e_x) = 0$

(5) $cov(e_z, e_y) = 0$

also yielding the estimand: $c = R_{zy \cdot x} = (R_{zy} - R_{zx}R_{yx})/(1 - R_{yx}^2)$,

**Assumption set $A_3$**

(1) $x = e_x$

(2) $y = bx + e_y$

(3) $z = cy + dx + e_z$

(4) $cov(e_z, e_x) = 0$

(7) $d = 0$

This set yields the instrumental-variable (IV) estimand: $c = R_{zx}/R_{yx}$.

Figure 2 provides a graphic illustration of these assumption sets, where each missing edge represents an assumption and each edge (i.e., an arrow or a bidirected arc) represents a relaxation of an assumption (since it permits the corresponding parameter to remain free). We see that $c$ is corroborated by three distinct set of assumptions, yielding two distinct estimands; the first two sets are degenerate, leading to the same estimand, hence $c$ is classified as 2-identified and 3-corroborated (see Definition 5).

Note that assumption (7), $d = 0$, is not needed for deriving $c = R_{zy \cdot x}$. Moreover, we cannot relax both assumption (4) and (6), as this would render $c$ non-identifiable. Finally, had we not separated (7) from (3), we would not be able to detect that $A_2$ is minimal, because it would appear as a superset of $A_3$.

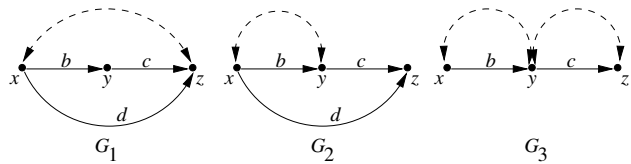

Figure 2: Graphs representing assumption sets $A_1, A_2$, and $A_3$, respectively.

It is also interesting to note that the natural estimand $c = R_{zy}$ is not selected as appropriate for $c$, because its derivation rests on the assumptions $\{(1), (2), (3), (4), (6), (7)\}$, which is a superset of each of $A_1, A_2$ and $A_3$. The implication is that $R_{zy}$ is not as robust to misspecification errors as the conditional regression coefficient $R_{zy \cdot x}$ or the instrumental variable estimand $R_{zx}/R_{yx}$. The conditional regression coefficient $R_{zy \cdot x}$ is robust to violation of assumptions (4) and (7) (see $G_1$ in Fig. 2) or assumptions (6) and (7) (see $G_2$ in Fig. 2), while the instrumental variable estimand $R_{zx}/R_{yx}$ is robust to violations of assumption (5) and (6), (see $G_3$, Fig. 2). The estimand $c = R_{zy}$, on the other hand, is robust to violation of assumption (6) alone, hence it is "dominated" by each of the other two estimands; there exists no data generating model that would render $c = R_{zy}$ unbiased and the $c = R_{zx}/R_{yx}$ (or $c = R_{zy \cdot x}$) biased. In contrast, there exist models in which $c = R_{zx}/R_{yx}$ (or $c = R_{zy \cdot x}$) is unbiased and $c = R_{zy}$ is biased; the graphs depicted in Fig. 2 represent in fact such models.

We now attend to the analysis of $b$. If we restrict the model to be recursive (i.e., feedback-less) and examine the set of assumptions embodied in the model of Fig. 1, we find that parameter $b$ is corroborated by only one minimal set of assumptions, given by:

(1) $x = e_x$

(2) $y = bx + e_y$

(6) $cov(e_x, e_y) = 0$

These assumptions yield the regression estimand, $b = R_{yx}$. Since any other derivation of $b$ must rest on these three assumptions, we conclude that no other set of assumptions can satisfy the minimality condition of Definition 4. Therefore, using Definition 6, $b$ is classified as just-identified.

Attempts to attribute to $b$ a second estimand, $b = R_{zx}/R_{zy}$, fail to recognize the fact that the second estimand is merely a noisy version of the first, for it relies on the same assumptions as the first, plus more. Therefore, if the two estimates of $b$ happen to disagree in a specific study, we can conclude that the disagreement must originates with violation of those extra assumptions that are needed for the second, and we can safely discard the second in favor of the first. Not so with $c$. If the two estimates of $c$ disagree, we have no reason to discard one in favor of the other, because the two rest on two distinct sets of assumptions, and it is always possible that either one of the two sets is valid. Conversely, if the two estimates of $c$ happen to coincide in a specific study, $c$ obtains a greater confirmation from the data since, for $c$ to be false, the coincidence of the two estimates can only be explained by an unlikely miracle. Not so with $b$. The coincidence of its two estimates might well be attributed to the validity of only those extra assumptions needed for the second estimate, but the basic common assumption needed for deriving $b$ (namely, assumption (6)) may well be violated.

## 5 GRAPHICAL TESTS FOR OVER-IDENTIFICATION

In this section we restrict our attention to parameters in the form of path coefficients, excluding variances and covariances of unmeasured variables, and we devise a graphical test for the over-identification of such parameters. The test rests on the following lemma, which generalizes Theorem 5.3.5 in [Pearl, 2000, p. 150], and embraces both instrumental variables and regression methods in one graphical criterion. (See also ibid, Definition 7.4.1, p. 248).

**Lemma 1** *(Graphical identification of direct effects) Let $c$ stand for the path coefficient assigned to the arrow $X \to Y$ in a causal graph $G$. Parameter $c$ is identified if there exists a pair $(W, Z)$, where $W$ is a node in $G$ and $Z$ is a (possibly empty) set of nodes in $G$, such that:*

1. *$Z$ consists of nondescendants of $Y$,*

2. *$Z$ d-separates $W$ from $Y$ in the graph $G_c$ formed by removing $X \to Y$ from $G$.*

3. *$W$ and $X$ are d-connected given $Z$, in $G_c$, or $W = X$.*

*Moreover, the estimand induced by the pair $(W, Z)$ is given by:*
$$c = \frac{cov(Y, W|Z)}{cov(X, W|Z)}.$$

The graphical test offered by Lemma 1 is sufficient but not necessary, that is, some parameters are identifiable, though no identifying $(W, Z)$ pair can be found in $G$ (see ibid, Fig. 5.11, p. 154). The test applies nevertheless to a large set of identification problems, and it can be improved to include several instrumental variables $W$. We now apply Lemma 1 to Definition 4, and associate the absence of a link with an "assumption."

**Definition 7** *(Maximal IV-pairs)*[3]
*A pair $(W, Z)$ is said to be an **IV-pair** for $X \to Y$, if it satisfies conditions (1–3) of Lemma 1. (IV connotes "instrumental variable.") An IV-pair $(W, Z)$ for $X \to Y$ is said to be maximal in $G$, if it is an IV-pair for $X \to Y$ in some graph $G'$ that contains $G$, and any edge-supergraph of $G'$ admits no IV-pair (for $X \to Y$), not even collectively.*[4]

**Theorem 1** *(Graphical test for over-identification)*
*A path parameter $c$ on arrow $X \to Y$ is over-identified if there exist two or more distinct maximal IV-pairs for $X \to Y$.*

**Corollary 1** *(Test for k-identifiability)*
*A path parameter $c$ on arrow $X \to Y$ is at least $k$-identified if there exist $k$ distinct maximal IV-pairs for $X \to Y$.*

**Example 2** *Consider the chain in Fig. 3(a). In this*

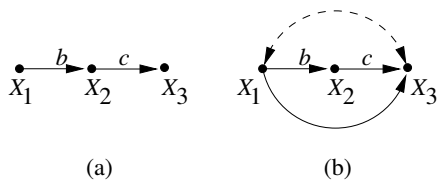

Figure 3:

*example, $c$ is 2-identified, because the pairs $(W = X_2, Z = X_1)$ and $(W = X_1, Z = 0)$ are maximal IV-pairs for $X_2 \to X_3$. The former yields the estimand $c = R_{32 \cdot 1}$, the latter yields $c = R_{31}/R_{21}$.*

Note that the robust estimand of $c$ is $R_{32 \cdot 1}$, not $R_{32}$. This is because the pair $(W = X_2, Z = \emptyset)$, which yields $R_{32}$, is not maximal; there exists an edge-supergraph of $G$ (shown in Fig. 3(b)) in which $Z = \emptyset$ fails to $d$-separate $X_2$ from $X_3$, while $Z = X_1$ does $d$-separate $X_2$ from $X_3$. The latter separation qualifies $(W = X_2, Z = X_1)$ as an IV-pair for $X_2 \to X_3$, and yields $c = R_{32 \cdot 1}$.

---

[3]Carlos Brito was instrumental in formulating this definition [Brito and Pearl, 2002ab].

[4]The qualification "not even collectively" aims to exclude graphs that admit no IV-pair for $X \to Y$, yet permit, nevertheless, the identification of $c$ through the collective action of $k$ IV-pairs for $k$ parameters (see [Pearl, 2000, Fig. 5.11] for examples). The precise graphical characterization of this class of graphs is currently under formulation, but will not be needed for the examples discussed in this paper.

The question remains how we can perform the test without constructing all possible supergraphs.

Every $(W, Z)$ pair has a set $S(W, Z)$ of maximally filled graphs, namely supergraphs of $G$ to which we cannot add any edge without spoiling condition (2) of Lemma 1. To test whether $(W, Z)$ leads to robust estimand, we need to test each member of $S(W, Z)$ so that no edge can be added without spoiling the identification of $c$. Thus, the complexity of the test rests on the size of $S(W, Z)$.

Graphs $G_1$ and $G_2$ in Fig. 2 constitute two maximally filled graphs for the IV-pair $(W = y, Z = x)$; $G_3$ is maximally filled for $(W = x, Z = \emptyset)$.

## 6 RELEVANCE-BASED FORMULATION

The preceding analysis shows ways of overcoming two major deficiencies in current methods of parameter estimation. The first, illustrated in Example 1, is the problem of *irrelevant over-identification*; certain assumptions in a model may render the model over-identified while playing no role whatsoever in the estimation of the parameters of interest. It is often the case that only selected portions of a model gather support through confrontation with the data, while others do not, and it is important to separate the former from the latter. The second is the problem of *irrelevant misspecifications*. If one or two of the model assumptions are incorrect, the model as a whole would be rejected as misspecified, though the incorrect assumptions may be totally irrelevant to the parameters of interest. For instance, if the assumption $cov(e_y, e_z) = 0$ in Example 1 (Figure 1) was incorrect, the constraint $R_{zx} = R_{yx} R_{zy}$ would clash with the data, and the model would be rejected, though the regression estimate $b = R_{yx}$ remains perfectly valid. The offending assumption in this case is irrelevant to the identification of the target quantity.

This section reformulates the notion of over-identification as a condition that renders a set of *relevant* assumptions (for a given quantity) testable.

If the target of analysis is a parameter $p$ (or a set of parameters), and if we wish to assess the degree of support that the estimation of $p$ earns through confrontation with the data, we need to assess the disparity between the data and the model assumptions, but we need to consider only those assumptions that are relevant to the identification of $p$, all other assumptions should be ignored. Thus, the basic notion needed for our analysis is that of "irrelevance"; when can we declare a certain assumption irrelevant to a given parameter $p$?

One simplistic definition would be to classify as relevant assumptions that are absolutely necessary for the identification of $p$. In the model of Figure 1, since $b$ can be identified even if we violate the assumptions $cov(e_z, e_y) = cov(e_z, e_x) = 0$, we declare these assumptions irrelevant to $b$, and we can ignore variable $z$ altogether. However, this definition would not work in general, because no assumption is absolutely necessary; any assumption can be disposed with if we enforce the model with additional assumptions. Taking again the model in Figure 1; the assumption $cov(e_y, e_z) = 0$ is not absolutely necessary for the identification of $c$, because $c$ can be identified even when $e_y$ and $e_z$ are correlated (see $G_3$ in Figure 2), yet we cannot label this assumption irrelevant to $c$.

The following definition provides a more refined characterization of irrelevance.

**Definition 8** *Let $A$ be an assumption embodied in model $M$, and $p$ a parameter in $M$. $A$ is said to be relevant to $p$ if and only if there exists a set of assumptions $S$ in $M$ such that $S$ and $A$ sustain the identification of $p$ but $S$ alone does not sustain such identification.*

**Theorem 2** *An assumption $A$ is relevant to $p$ if and only if $A$ is a member of a minimal set of assumptions sufficient for identifying $p$.*

**Proof:**
Let $msa$ abbreviate "minimal set of assumptions sufficient for identifying $p$" and let the symbol "$\models p$" denote the relation "sufficient for identifying $p$" ($\not\models p$, its negation). If $A$ is a member of some $msa$ then, by definition, it is relevant. Conversely, if $A$ is relevant, we will construct a $msa$ of which $A$ is a member. If $A$ is relevant to $p$, then there exists a set $S$ such that $S + A \models p$ and $S \not\models p$. Consider any minimal subset $S'$ of $S$ that satisfies the properties above, namely

$$S' + A \models p \text{ and } S' \not\models p,$$

and, for every proper subset $S''$ of $S'$, we have (from minimality)

$$S'' + A \not\models p \text{ and } S'' \not\models p,$$

(we use monotonicity here; removing assumptions cannot entail any conclusion that is not entailed before removal). The three properties: $S' + A \models p$, $S' \not\models p$, and $S'' + A \not\models p$ (for all $S'' \subset S'$) qualify $S' + A$ as $msa$, and completes the proof of Theorem 2. QED.

Thus, if we wish to prune from $M$ all assumptions that are irrelevant to $p$, we ought to retain only the *union* of all minimal sets of assumptions sufficient for identifying $p$. This union constitutes another model, in which all assumptions are relevant to $p$. We call this new model the *p-relevant* submodel of $M$, $M_p$ which we formalize by a definition.

**Definition 9** *Let $A_M$ be the set of assumptions embodied in model $M$, and let $p$ be an identifiable parameter in $M$. The p-relevant submodel of $M$, denoted $M_p$ is a model consisting of the union of all minimal subsets of $A_M$ sufficient for identifying $p$.*

We can naturally generalize this definition to any quantity of interest, not necessarily a single parameter.

**Definition 10** *Let $A_M$ be the set of assumptions embodied in model $M$, and let $q$ be any quantity identifiable in $M$. The q-relevant submodel of $M$, denoted $M_q$ is a model consisting of the union of all minimal subsets of $A_M$ sufficient for identifying $q$.*

We can now associate with any quantity $q$ in a model properties that are normally associated with models, for example, fit indices, degree of fitness, degrees of freedom ($df$) and so on; we simply compute these properties for $M_q$, and attribute the results to $q$. For example, if $D_q$ measures the fitness of $M_q$ to a body of data, we can say that quantity $q$ has disparity $D_q$ with $df(q)$ degrees of freedom.

Consider the model of Figure 1. If $q = b, M_b$ would consist of one assumption, $cov(e_x, e_y) = 0$, since this assumption is minimally sufficient for the identification of $b$. Discarding all other assumptions of $A$ is equivalent to considering the arrow $x \rightarrow y$ alone, while discarding the portions of the model associated with $z$. Since $M_b$ is saturated (that is, just identified) it has zero degrees of freedom, and we can say that $b$ has zero degrees of freedom, or $df(b) = 0$. If $q = c$, $M_c$ would be the entire model $M$, because the union of assumption sets $A_1, A_2$ and $A_3$ span all the seven assumptions of $M$. We can therefore say that $c$ has one degree of freedom, or $df(c) = 1$. This means that the claim $c = c_0$ constrains the covariance matrix by a one-dimensional manifold.

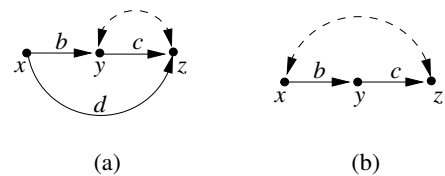

Figure 4:

Now assume that the quantity of interest, $q$, stands for the total effects of $x$ on $z$, denoted $TE(x, z)$. There are two minimal subsets of assumptions in $M$ that are sufficient for identifying $q$. Figure 4 represents these subsets through their respective (maximal) subgraphs;

model 4(a) yields the estimand $TE(x,z) = R_{zx}$, while 4(b) yields $TE(x,z) = R_{yx}R_{zy\cdot x}$. Note that although $c$ is not identified in the model of Figure 4(a), the total effect of $x$ of on $z$, $TE(x,z) = d + bc$, is nevertheless identified. The union of the two assumption sets coincides with the original model $M$ (as can be seen by taking the intersection of the corresponding arcs in the two subgraphs. Thus, $M = M_q$, and we conclude that $TE(x,z)$ is 2-identified, and has one degree of freedom.

For all three quantities, $b$, $c$ and $TE(x,z)$ we obtained degrees of freedom that are one less than the corresponding degrees of identification, $k(q) = df(q)$. This is a general relationship, as shown in the next Theorem.

**Theorem 3** *The degrees of freedom associated with any quantity $q$ computable from model $M$ is given by $df(q) = k(q) - 1$, where $k(q)$ stands for the degree of identifiability (Definition 5).*

**Proof**
$df(q)$ is given by the number of independent equality constraints that model $M_q$ imposes on the covariance matrix. $M_q$ consists of $m$ distinct $msa$'s, which yield $m$ estimands for $q$, $q = q_i(\sigma)$, $i = 1, \ldots, m$, $k$ of which are distinct. Since all these $k$ functions must yield the same value for $q$, they induce $k - 1$ independent equality constraints:

$$q_i(\sigma) = q_{i+1}(\sigma), \ i = 1, 2, \ldots, k-1$$

This amounts to $k-1$ degrees of freedom for $M_q$, hence, $df(q) = k(q) - 1$.  QED.

We thus obtain another interpretation of $k$, the degree of identifiability; $k$ equals one plus the degrees of freedom associated with the $q$-relevant submodel of $M$.

## 7 CONCLUSIONS

This paper gives a formal definition to the notion of robustness, or over-identification of causal parameters. This definition resolves long standing difficulties in rendering the notion of robustness operational. We also established a graphical method of quantifying the degree of robustness. The method requires the construction of maximal supergraphs sufficient for rendering a parameter identifiable and counting the number of such supergraphs with distinct estimands.

The qualitative approach of this paper assumes that all modeling assumptions have equal weight, and does not account for the case where a modeler can express different degrees of belief in the validity of the various assumptions. A Bayesian approach would be natural for incorporating this extra knowledge, when available, but would encounter the problem of computing the posterior probability of the causal claim, integrated over all assumption sets that have a non-zero prior probability.

### Acknowledgment

This research was partially supported by AFOSR grant #F49620-01-1-0055, NSF grant #IIS-0097082, and ONR (MURI) grant #N00014-00-1-0617.

### References

[Bollen, 1989] K.A. Bollen. *Structural Equations with Latent Variables.* John Wiley, New York, 1989.

[Bowden and Turkington, 1984] R.J. Bowden and D.A. Turkington. *Instrumental Variables.* Cambridge University Press, Cambridge, England, 1984.

[Brito and Pearl, 2002a] C. Brito and J Pearl. Generalized instrumental variables. In A. Darwiche and N. Friedman, editors, *Uncertainty in Artificial Intelligence, Proceedings of the Eighteenth Conference*, pages 85–93. Morgan Kaufmann, San Francisco, 2002.

[Brito and Pearl, 2002b] C. Brito and J Pearl. A graphical criterion for the identification of causal effects in linear models. In *Proceedings of the Eighteenth National Conference on Artificial Intelligence*, pages 533–538. AAAI Press/The MIT Press, Menlo Park, CA, 2002.

[Joreskog, 1979] K.G. Joreskog. *Advances in Factor Analysis and Structural Equation Models*, chapter Structural equation models in the social sciences: Specification, estimation and testing, pages 105–127. Abt Books, Cambridge, MA, 1979.

[MacCallum, 1995] R.C. MacCallum. Model specification, procedures, strategies, and related issues (chapter 2). In R.H. Hoyle, editor, *Structural Equation Modeling*, pages 16–36. Sage Publications, Thousand Oaks, CA, 1995.

[Pearl, 2000] J. Pearl. *Causality: Models, Reasoning, and Inference.* Cambridge University Press, New York, 2000.